\pdfoutput=1

\documentclass[11pt]{article}

\usepackage{acl}

\usepackage{times}
\usepackage{latexsym}

\usepackage[T1]{fontenc}

\usepackage[utf8]{inputenc}

\usepackage{microtype}

\usepackage{amsmath,amsfonts,bm}
\usepackage{booktabs} %
\usepackage{url}
\usepackage{graphicx}
\usepackage{array}
\usepackage{color}
\usepackage{makecell}
\usepackage{multirow}
\usepackage{xspace}

\newcommand{\eg}{\textit{e.g.}} %
\newcommand{\start}[1]{\vspace{.8mm}\noindent{{\bf #1}.}}

\newcommand{\upv}{\vspace{-.0cm}}
\newcommand{\downv}{\vspace{-.1cm}}

\newcommand{\base}{BERTSum\xspace}

\definecolor{gred}{RGB}{219,68,55}
\definecolor{gblue}{RGB}{66,133,244}
\definecolor{gyellow}{RGB}{244,180,0}
\definecolor{ggreen}{RGB}{15,157,88}
\definecolor{ggrey}{RGB}{115,115,115}

\newcommand{\colorR}[1]{\textcolor{gred}{\textbf{#1}}}
\newcommand{\colorG}[1]{\textcolor{ggreen}{\textbf{#1}}}

\title{Constrained Abstractive Summarization:\\ Preserving Factual Consistency with Constrained Generation}

\author{Yuning Mao$^{1}$, Xiang Ren$^2$, Heng Ji$^1$, Jiawei Han$^1$ \\
$^1$University of Illinois, Urbana-Champaign \quad
$^2$University of Southern California\\
$^1$\{yuningm2, hengji, hanj\}@illinois.edu $\quad$ $^2$xiangren@usc.edu
}
\date{}

\begin{document}
\maketitle

\begin{abstract}
Despite significant progress, state-of-the-art abstractive summarization methods are still prone to hallucinate content inconsistent with the source document.
In this paper, we propose Constrained Abstractive Summarization (CAS), a general setup that preserves the factual consistency of abstractive summarization by specifying tokens as constraints that must be present in the summary.
We adopt lexically constrained decoding, a technique generally applicable to autoregressive generative models, to fulfill CAS and conduct experiments in two scenarios: (1) automatic summarization without human involvement, where keyphrases are extracted from the source document and used as constraints; (2) human-guided interactive summarization, where human feedback in the form of manual constraints are used to guide summary generation.
Automatic and human evaluations on two benchmark datasets demonstrate that CAS improves both lexical overlap (ROUGE) and factual consistency of abstractive summarization.
In particular, we observe up to 13.8 ROUGE-2 gains when only \textit{one} manual constraint is used in interactive summarization.\footnote{\small Our code can be found at \url{https://github.com/morningmoni/EDE}.}
\end{abstract}

\section{Introduction}

Although abstractive summarization has achieved significant progress with advances in seq2seq learning \cite{see-etal-2017-get} and language model pre-training \cite{liu-lapata-2019-text,lewis2019bart}, its generation process is typically unconstrained: abstractive models learn to generate summaries in a completely data-driven manner using document-summary pairs.
As a consequence, they are prone to hallucinate content not entailed by the source documents (\eg, producing unseen entities or unfaithful facts)~\cite{kryscinski-etal-2020-evaluating,maynez-etal-2020-faithfulness}.
Such \textit{factual inconsistencies} in the summary hinder the practicability of abstractive models in real-world applications.

\begin{table}[t]
        \resizebox{1.0\columnwidth}{!}{
        \begin{tabular}{p{8cm}}
            \toprule
                \textbf{Reference}: sir tom jones is to return as one of the judges on talent show \colorG{the voice uk} when it moves to \colorG{itv} next year. \\
             \textbf{Unconstrained}: pop star sir tom jones is to return to the \colorR{bbc} 's \colorR{voice uk} after a two-year absence. \\
             \textbf{Constrained}: singer sir tom jones is to return to \colorG{itv} 's \colorG{the voice uk} next year after a two-year absence.\\
             \midrule
            \textbf{Reference}: syrian refugees facing their first \colorG{christmas} in wales are sure to get a `` warm welsh welcome '', the first minister has said. \\
             \textbf{Unconstrained}: the first minister has said wales is \colorR{`` more important than ever '' in the new year}. \\
             \textbf{Constrained}: the first welsh councils to welcome refugees in the uk this \colorG{christmas} have been praised by the first minister.\\
             
             \midrule
             
            \textbf{Reference}: a four-month consultation which could help decide the location of the uk 's first spaceport ends on monday with one site in \colorG{gwynedd} being considered. \\
             \textbf{Unconstrained}: plans for a uk spaceport and spaceport in snowdonia have been backed by \colorR{a council}. \\
             \textbf{Constrained}: plans to build a uk spaceport in snowdonia have been backed by \colorG{gwynedd} council.\\

            \bottomrule
        \end{tabular}
        }
    \upv
    \caption{\textbf{Unconstrained and constrained summaries of the same model without additional training}. \colorG{Constrained} (\colorR{replaced}) tokens are in green (red).}
    \label{table_cases}
    \downv
    \end{table}

Recent studies~\cite{matsumaru-etal-2020-improving,zhu-etal-2021-enhancing,dong-etal-2020-multi} on preserving the factual consistency of abstractive summarization are often highly coupled with specific models or incur inferior lexical overlap (lower ROUGE).
In this paper, we propose Constrained Abstractive Summarization (CAS), a more general setup where a set of tokens are used as constraints and required to be present in the summary. We show that CAS improves lexical overlap and factual consistency of abstractive summarization simultaneously.

We consider two scenarios for CAS: (1) automatic summarization without human involvement, where we extract keyphrases from the source document and use them as constraints;
(2) human-guided interactive summarization, where human feedback in the form of manual constraints  are used to guide CAS towards human preferences.
We enforce the constraints by lexically constrained decoding~\cite{post-vilar-2018-fast}, which only functions during inference and can be easily integrated into different autoregressive abstractive models with beam search decoding.
In this way, one can conduct CAS on (almost) any fine-tuned models without (often expensive) re-training.

CAS preserves the factual consistency of abstractive summarization in two ways according to our observations.
First, the added constraints can often replace their unfaithful counterparts in the unconstrained summary (produced by the same model) and help reduce model hallucination. For example, in Table~\ref{table_cases}, when given ``\textit{ITV}'' as a constraint the model corrects ``\textit{BBC}'' to ``\textit{ITV}'' as ``\textit{The Voice UK}'' is acquired by ``\textit{ITV}''.
Second, when adding important entities not found in the unconstrained summary as constraints, the model is more likely to generate summaries that are focused on these factual entities (``\textit{Christmas}'') and more specific (``\textit{a council}'' changed to ``\textit{Gwynedd council}'').

We study CAS on the CNN/Daily Mail (CNNDM) \cite{nallapati-etal-2016-abstractive-new} and XSum \cite{narayan-etal-2018-dont} datasets with \base \cite{liu-lapata-2019-text} as an example of the base model.
Automatic and human evaluations show that CAS improves abstractive summarization on both lexical overlap (ROUGE) and factual consistency.
To our knowledge, CAS is the first method to improve both aspects simultaneously and consistently.
Moreover, \base under CAS achieves better performance than \textit{more expensive} methods such as BART \cite{lewis2019bart} and PEGASUS \cite{zhang2019pegasus} by simply using \textit{one} manual constraint during inference, which demonstrates the benefits of CAS in interactive summarization and shows its great potential in future development.

\section{Method}

\subsection{Task Formulation}

We define a constraint set $\mathcal{C} = \{c_1, c_2, ... c_N \}$ as a set of text spans of arbitrary length.
Given document-reference pair $(d, r)$ and abstractive model $\mathcal{M}$, Constrained Abstractive Summarization (CAS) generates a summary $s$ for document $d$ using $\mathcal{M}$ with the presence of all the text spans in $\mathcal{C}$.
We denote an unconstrained summary generated by the same $\mathcal{M}$ (with $\mathcal{C} = \emptyset$) as $s'$.
We aim to create a constraint set $\mathcal{C}$ that has a high overlap with $r$ to ensure its quality and low overlap with $s'$ to bring additional information, such that CAS with $\mathcal{C}$ improves the quality of the generated summary.

\subsection{Constraint Creation}
CAS is useful for improving abstractive summarization, especially on factual consistency in two scenarios.
First, one can create the constraints automatically without human involvement, where the keyphrases in the source document are natural choices for constraints to preserve factual consistency (Sec. \ref{sec:method_noisy}).
Second, for interactive summarization, when an automatic summary contains factual errors or lacks certain information, a human editor can manually add the corrected or missing facts as constraints during post-editing (Sec. \ref{sec:method_gold}).

\subsubsection{Automatic Constraints}
\label{sec:method_noisy}

To obtain constraints automatically, we adopt a state-of-the-art supervised keyphrase extraction method, BERT-KPE~\cite{sun2020joint}, to extract keyphrases from the source document, as we find commonly used unsupervised methods \cite{campos2020yake} insufficient to provide high-quality constraints.
Similar to recent studies \cite{nan-etal-2021-entity,narayan2021planning}, we focus on the factual consistency of entities and noun phrases.
We use spaCy~\cite{spacy2} to find named entities and noun phrases in the reference summaries of the training set, and treat those appearing in the source documents as positive training examples.
During test time, we exclude the extracted keyphrases appearing in $s'$ such that only constraints bringing additional information are used.

\subsubsection{Manual Constraints}
\label{sec:method_gold}

As automatic summarization is still imperfect and users also have different preferences or information needs, human-guided interactive summarization has gained increasing popularity \cite{avinesh2018sherlock,gao-etal-2018-april,shapira2020evaluating}.
Since human feedback is expensive to obtain, we simulate it by taking tokens in the reference summary as manual constraints.
For example, a user may revise a summary by adding entities they deem important but not in the system summary. We simulate such edits by taking entities in the reference but not system summary as constraints.
Similar simulations with the use of the references have been widely adopted in interactive machine translation~\cite{cheng-etal-2016-primt,hokamp-liu-2017-lexically,post-vilar-2018-fast,chen2020lexical} but not yet explored in summarization.
CAS with manual constraints is useful for interactive summarization where users do not have to (re)write the entire summary but provide minimal guidance, and also serves as an upper bound for automatic summarization.

\subsection{Lexically Constrained Decoding}

There are many different means to fulfill CAS as its formulation is general.
Here, we explore the effectiveness of using lexically constrained decoding, namely, dynamic beam allocation (DBA)~\cite{post-vilar-2018-fast}.
At a high level, DBA divides the beam during beam search to store hypotheses satisfying different numbers of constraints and adds unmet constraints at each decoding step.
DBA ensures the presence of constraints by allowing the EOS token only when all the constraints are met.
We choose DBA due to its faster speed than other counterparts~\cite{hokamp-liu-2017-lexically} -- a complexity of $\mathcal{O}(1)$ in the number of constraints.
Moreover, DBA completely functions during inference and can be easily incorporated into different models for the evaluation under CAS, which is preferable over methods that modify the training process \cite{zhang-etal-2020-pointer}, since we want to keep the carefully designed, often expensive abstractive models intact rather than re-train them with a different objective.

\section{Experiments}

\section{Experiment Setup}

We conduct experiments on the CNNDM \cite{nallapati-etal-2016-abstractive-new} and XSum \cite{narayan-etal-2018-dont} datasets.
We use one state-of-the-art abstractive model, \base \cite{liu-lapata-2019-text}, as our base model $\mathcal{M}$.
For automatic constraints, all N-grams up to $N=5$ are considered as constraint candidates.
To reduce noise, we take top $k$ keyphrases with scores greater than $v$ extracted by BERT-KPE as the constraints, which is tuned on the validation set with $k=3$ and $v=1.6$. 
We set beam size to 10 when automatic constraints are used and 5 for manual constraints unless otherwise specified.
For automatic evaluation on factual consistency, we do not conduct data filtering and evaluate on the full dataset.

\subsection{CAS for Automatic Summarization}
\label{sec:exp_noisy}

\start{Constraint Creation}
For constraint creation, we achieve 0.76 Prec@1 / 0.49 F1@5 on CNNDM, and 0.67 Prec@1 / 0.40 F1@5 on XSum, which suggests that the automatically extracted constraints are of reasonable quality and ready to use.

\start{Lexical Overlap}
Table~\ref{tab_noisy} shows the comparison of our base model and CAS in ROUGE.\footnote{Our results are slightly different from \citet{liu-lapata-2019-text} despite using its official code and model weights.}
CAS improves \base on both datasets \textit{with statistical significance}, which is achieved without any model training but enforcing the constraints.
As will be shown in Sec.~\ref{sec:exp_gold}, larger gains can be achieved when the keyphrase extraction module is improved.

\begin{table}[ht]
\centering

\scalebox{.83}{
\begin{tabular}{cl ccccc}
\cmidrule[0.06em]{2-7}
&\textbf{Method} & \textbf{R-1} & \textbf{R-2} & \textbf{R-L} & \textbf{Ent-F1} & \textbf{Sup} \\
\cmidrule{2-7}
\multirow{2}{*}{\scriptsize \rotatebox[origin=c]{90}{CNNDM}}&\base &42.00	&19.44	&38.98 & 36.3 & 89.1 \\
&\quad\ +CAS &\textbf{42.48} &\textbf{19.56} &\textbf{39.43} & \textbf{37.0} & \textbf{94.0} \\ %
\cmidrule{2-7}
\multirow{2}{*}{\footnotesize \rotatebox[origin=c]{90}{XSum}}&\base & 38.91	&16.54	&31.30 & 31.2 & 72.3\\
&\quad\ +CAS  &\textbf{39.19} &\textbf{16.75} &\textbf{31.56} & \textbf{35.1} & \textbf{73.0}\\ %

\cmidrule[0.06em]{2-7}
\end{tabular}
}
\upv
\caption{\textbf{CAS with automatically extracted keyphrases}. Improvements are statistically significant ($p < 0.05$) under approximate randomization test and  paired bootstrap resampling test.}
\label{tab_noisy}
\downv
\end{table}

\start{Factual Consistency}
More importantly, we observe consistent gains of CAS in factual consistency. In automatic evaluation, CAS improves \base by 0.7/3.9 on entity-level F1 \cite{nan-etal-2021-entity} and 4.9/0.7 on support score \cite{matsumaru-etal-2020-improving} for CNNDM/XSum, respectively.
In human evaluation, we randomly sample 50 examples and ask three expert annotators to compare the quality of the two systems.
Human ratings show that 34\% constrained summaries achieve better factual consistency, 52\% are similar to their unconstrained counterparts, and only 14\% samples become worse, confirming that CAS better preserves factual consistency while achieving higher ROUGE. 
To our knowledge, CAS is the first approach to improve on both aspects.
More details and 10 examples with analysis covering both good and bad cases are provided in App.~\ref{sec:human} and~\ref{sec:examples}.

\subsection{CAS for Interactive Summarization}
\label{sec:exp_gold}

In Table \ref{tab_xsum_hardDec}, we show the results of CAS in interactive summarization where various manual constraints simulated by the reference summary (ref) are used: \textit{Entity} (named entities in ref), \textit{NP} (noun phrases in ref), \textit{Random-4} (4 random tokens in ref), \textit{Phrase-4} (4 continuous tokens in ref), \textit{miss} (tokens not found in the unconstrained summary), and \textit{src} (tokens must appear in the source document).

\begin{table}[ht]
\centering

\resizebox{\columnwidth}{!}{
\scalebox{.99}{
\begin{tabular}{lcccc}
\toprule
\multirow{2}{*}{\textbf{Constraint Type}}& \multicolumn{4}{c}{\textbf{XSum}}\\
 & \textbf{R-1} & \textbf{R-2} & \textbf{R-L}  & $\bar{|\mathcal{C}|}$\\
 
\midrule
None &38.91	&16.54	&31.30 & -\\
Entity                  &         46.88  &         20.21  &         33.09  & \textbf{5.77} \\
Entity $\wedge$ miss            &         46.87  &         21.55  &         33.92  &         4.01  \\
Random-4                &         46.42  &         17.73  &         33.41  &         4.42  \\
Random-4 $\wedge$ miss          & \textbf{52.09} &         20.71  &         35.96  &         4.36  \\
Entity $\wedge$ miss $\wedge$ src     &         40.79  &         17.31  &         32.07  &         0.80  \\
Entity+NP$ \wedge$ miss $\wedge$ src &         43.30  &         18.82  &         33.07  &         2.02  \\
Phrase-4                 &         47.56  & 27.71 & 38.72 &         4.41  \\
Phrase-4 (beam size 50)                 &         49.76  & \textbf{30.33} & \textbf{42.55} &         4.41  \\
\midrule
BART {\small \cite{lewis2019bart}}	&45.14	&22.27	&37.25 &-\\
PEGASUS {\small \cite{zhang2019pegasus}}	&47.21	&24.56	&39.25 &-\\

\bottomrule
\end{tabular}
}
}
\upv
\caption{\textbf{Comparison of various types of manual constraints}. $\bar{|\mathcal{C}|}$ denotes the averaged total number of tokens in the constraint set $\mathcal{C}$. $\wedge$ denotes AND operation.
}
\label{tab_xsum_hardDec}
\downv
\end{table}

There are many interesting findings when applying CAS to interactive summarization.
For example, adding all constraints improves less than only adding missing ones ($\wedge$ miss), possibly because the model can already generate those tokens but still wastes some beams to store such constraints and thus fails to search for better alternatives;
Using random constraints (Random-4) sometimes incurs inarticulate output in our manual examination;
The improvement is not as significant when requiring the presence of constraints in the source document ($\wedge$ src), which coincides with recent findings that reference summaries often involve extrinsic information \cite{maynez-etal-2020-faithfulness};
Finally, by using only one phrase as guidance during inference without additional training (Phrase-4), CAS improves \base up to 13.8 in ROUGE-2, outperforming state-of-the-art methods.

In Table \ref{tab_cnndm_gold}, we list the performance of CAS when manual constraints are provided on CNNDM.
The performance gains on CNNDM are not as significant as on XSum, possibly because the summaries in CNNDM are much longer and we use a similar number of constraints on both datasets.
Nevertheless, CAS still helps \base achieve state-of-the-art performance by only using \textit{one} phrase as manual constraint.

\begin{table}[ht]
\centering

\resizebox{\columnwidth}{!}{
\scalebox{.98}{
\begin{tabular}{lrrr}
\toprule
\multirow{2}{*}{\textbf{Constraint Type}}& \multicolumn{3}{c}{\textbf{CNNDM}}\\
 & \textbf{R-1} & \textbf{R-2} & \textbf{R-L} \\
\midrule
None &42.00	&19.44	&38.98 \\
Entity &43.31 &19.57 &40.05 \\
Phrase-4 &         43.37  &         21.57  &         40.82  \\
Phrase-4 (beam size 20)  &         \textbf{45.14}  &         \textbf{23.21}  &         \textbf{42.43}   \\
\midrule
BART \cite{lewis2019bart}	&44.16	&21.28	&40.90\\
PEGASUS \cite{zhang2019pegasus}	&44.17	&21.47	&41.11\\

\bottomrule
\end{tabular}
}
}
\upv
\caption{\textbf{Performance of CAS when manual constraints are provided on CNNDM}.}
\label{tab_cnndm_gold}
\downv
\end{table}

\subsection{Analysis and Discussion}

\begin{figure}[ht]
    \centering
    \includegraphics[width=0.86\linewidth]{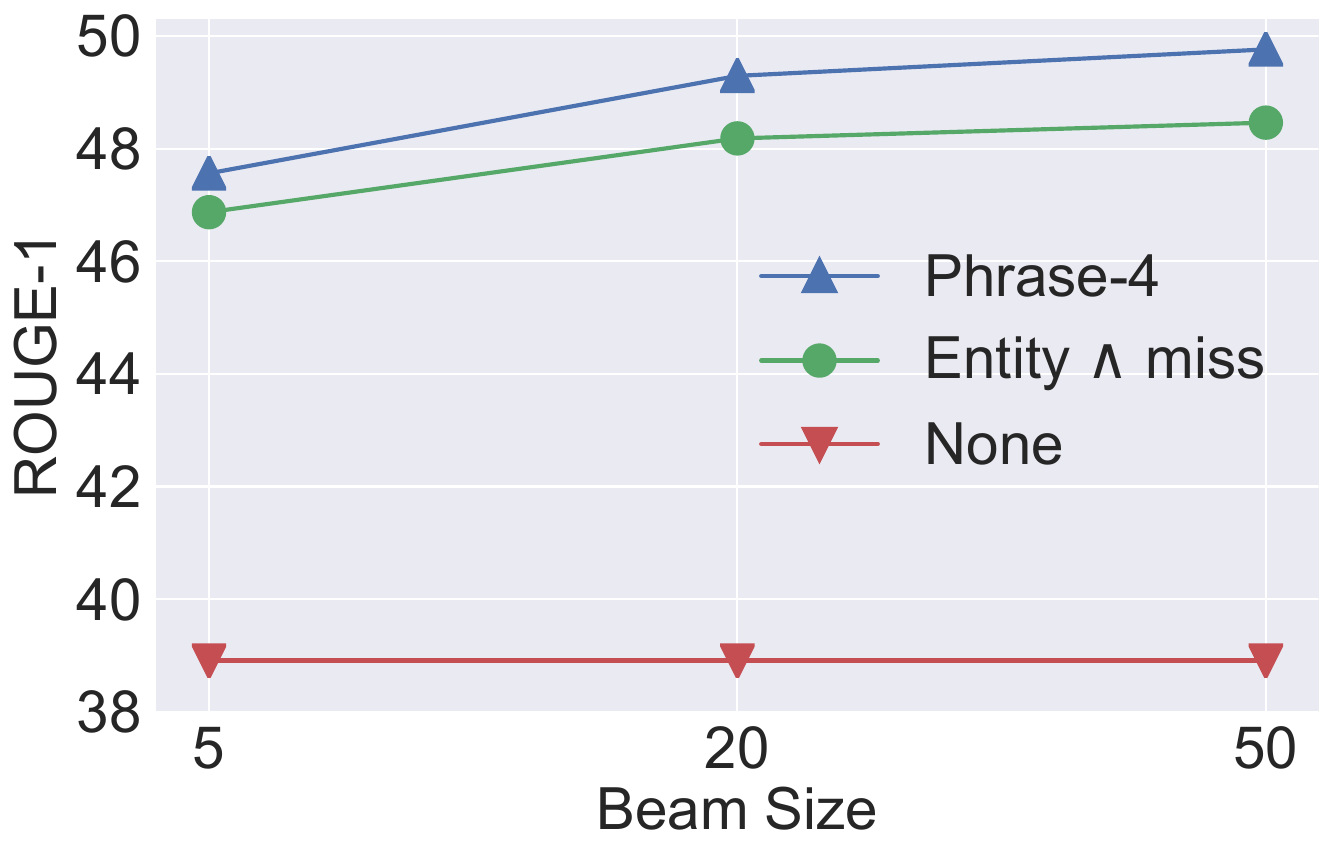}
    \upv
    \caption{\textbf{Performance changes by beam size on XSum}.  Larger beam leads to better results for CAS.}
    \label{fig:beam}
    \downv
\end{figure}

\start{CAS guides generation}
As shown in Fig.~\ref{fig:beam}, using a larger beam size consistently leads to better performance for CAS with the same constraints, while the performance change is negligible for unconstrained summarization (detailed numbers in App.~\ref{app:beam}).
Such results imply that the gains of CAS are based on a guided generation process that exploits model potential, rather than merely access to a few manual constraints.

\start{CAS is not random insertion}
We observe that CAS typically inserts constraints at proper positions and often corrects unfaithful information (Table~\ref{table_cases} and more examples in App.~\ref{sec:examples}).
To further study the gap between CAS and random insertion, we directly append the constraints to the end of unconstrained summaries as a baseline.
As listed in Table~\ref{tab_random}, simply appending constrained tokens cannot obtain the same performance gains as CAS and apparently leads to nonfluency too.

\begin{table}[ht]
\centering

\scalebox{.86}{
\begin{tabular}{lccc}
\toprule
\textbf{Method (beam size 50)} & \textbf{R-1} & \textbf{R-2} & \textbf{R-L} \\
\midrule
Random (Entity $\wedge$ miss) &47.03	&20.79	&34.28 \\
CAS \quad \ \   (Entity $\wedge$ miss) &         \textbf{48.46}  &         \textbf{24.17}  &         \textbf{37.47} \\
\midrule
Random (Phrase-4) &         46.71   &         26.11  &         35.69  \\
CAS \quad \ \ (Phrase-4)  & \textbf{49.76} & \textbf{30.33} & \textbf{42.55}   \\

\bottomrule
\end{tabular}
}
\upv
\caption{\textbf{CAS outperforms random insertion by a large margin.} Results are on the XSum dataset.}
\label{tab_random}
\downv

\end{table}

\start{Runtime overhead is low}
For automatic constraints, since keyphrase extraction is independent of summarization, one can use the same keyphrases on different summarization models and the total time is amortized.
For constrained decoding, a detailed runtime comparison between unconstrained and constrained generation is shown in Table~\ref{tab_time}.
The overhead of constrained generation is acceptable when the beam size is not large (we use a beam size of 5/10 for most experiments).
Moreover, a faster DBA implementation \cite{hu2019improved} would further reduce the runtime.

\begin{table}[ht]
\centering

\scalebox{.89}{
\begin{tabular}{lcc}
\toprule
\textbf{Beam Size} & \textbf{Unconstrained} & \textbf{CAS} \\
\midrule
5 & 25min	& 33min\\
10 & 44min & 68min  \\
20 & 70min	& 4h\\
50 & 2.5h & 10h\\

\bottomrule
\end{tabular}
}

\upv
\caption{\textbf{Comparison of inference time} for the XSum test set on one GTX 1080 Ti GPU.}
\label{tab_time}
\downv
\end{table}

\section{Related Work}

\start{Factual Consistency of Summarization}
As ROUGE \cite{lin-2004-rouge} does not correlate well with factual consistency \cite{falke-etal-2019-ranking},
model-based metrics \cite{goodrich2019assessing,kryscinski-etal-2020-evaluating} are proposed to measure factual consistency explicitly.
However, their performance is unsatisfactory and to date, there is still no commonly accepted metric beyond human evaluation.
Prior studies \cite{matsumaru-etal-2020-improving,zhu-etal-2021-enhancing,dong-etal-2020-multi} generally improve factual consistency at the expense of ROUGE while CAS improves on both aspects simultaneously.

\smallskip
\start{Constrained Generation}
Constrained Generation is useful in various tasks such as machine translation \cite{hokamp-liu-2017-lexically} and data augmentation \cite{hu-etal-2019-improved}.
Although approaches like copy mechanism \cite{see-etal-2017-get} that encourage models to copy words from source documents have been widely adopted, they are often insufficient to reduce model hallucination \cite{maynez-etal-2020-faithfulness}.
To our knowledge, constrained summarization like CAS that requires certain tokens must be present in a summary is unexplored.

\section{Conclusion}
In this paper, we propose constrained abstractive summarization (CAS), a general setup that can be easily incorporated into existing abstractive models to preserve factual consistency.
We demonstrate that CAS leads to higher-quality, especially more factually consistent summaries, in automatic and interactive summarization under both automatic and human evaluations.
For future work, we will explore and facilitate alternative means beyond lexically constrained decoding to fulfill CAS.

\section*{Acknowledgments}
We thank Junxian He, Mike Lewis, Yanru Qu, and Jiaming Shen for helpful discussions.

\bibliography{CAS,anthology}

\begin{thebibliography}{29}
\expandafter\ifx\csname natexlab\endcsname\relax\def\natexlab#1{#1}\fi

\bibitem[{Avinesh et~al.(2018)Avinesh, Binnig, H{\"a}ttasch, Meyer, and
  {\"O}zyurt}]{avinesh2018sherlock}
PVS Avinesh, Carsten Binnig, Benjamin H{\"a}ttasch, Christian~M Meyer, and
  Orkan {\"O}zyurt. 2018.
\newblock Sherlock: A system for interactive summarization of large text
  collections.
\newblock \emph{Proc. VLDB Endow.}, 11(12):1902--1905.

\bibitem[{Campos et~al.(2020)Campos, Mangaravite, Pasquali, Jorge, Nunes, and
  Jatowt}]{campos2020yake}
Ricardo Campos, V{\'\i}tor Mangaravite, Arian Pasquali, Al{\'\i}pio Jorge,
  C{\'e}lia Nunes, and Adam Jatowt. 2020.
\newblock Yake! keyword extraction from single documents using multiple local
  features.
\newblock \emph{Information Sciences}, 509:257--289.

\bibitem[{Chen et~al.(2020)Chen, Chen, Wang, and Li}]{chen2020lexical}
Guanhua Chen, Yun Chen, Yong Wang, and Victor~OK Li. 2020.
\newblock Lexical-constraint-aware neural machine translation via data
  augmentation.
\newblock In \emph{Proceedings of the Twenty-Ninth International Joint
  Conference on Artificial Intelligence, IJCAI-20}, pages 3587--3593.

\bibitem[{Cheng et~al.(2016)Cheng, Huang, Chen, Dai, and
  Chen}]{cheng-etal-2016-primt}
Shanbo Cheng, Shujian Huang, Huadong Chen, Xin-Yu Dai, and Jiajun Chen. 2016.
\newblock \href {https://doi.org/10.18653/v1/N16-1148} {{PRIMT}: A pick-revise
  framework for interactive machine translation}.
\newblock In \emph{Proceedings of the 2016 Conference of the North {A}merican
  Chapter of the Association for Computational Linguistics: Human Language
  Technologies}, pages 1240--1249, San Diego, California. Association for
  Computational Linguistics.

\bibitem[{Dong et~al.(2020)Dong, Wang, Gan, Cheng, Cheung, and
  Liu}]{dong-etal-2020-multi}
Yue Dong, Shuohang Wang, Zhe Gan, Yu~Cheng, Jackie Chi~Kit Cheung, and Jingjing
  Liu. 2020.
\newblock \href {https://doi.org/10.18653/v1/2020.emnlp-main.749} {Multi-fact
  correction in abstractive text summarization}.
\newblock In \emph{Proceedings of the 2020 Conference on Empirical Methods in
  Natural Language Processing (EMNLP)}, pages 9320--9331, Online. Association
  for Computational Linguistics.

\bibitem[{Falke et~al.(2019)Falke, Ribeiro, Utama, Dagan, and
  Gurevych}]{falke-etal-2019-ranking}
Tobias Falke, Leonardo F.~R. Ribeiro, Prasetya~Ajie Utama, Ido Dagan, and Iryna
  Gurevych. 2019.
\newblock \href {https://doi.org/10.18653/v1/P19-1213} {Ranking generated
  summaries by correctness: An interesting but challenging application for
  natural language inference}.
\newblock In \emph{Proceedings of the 57th Annual Meeting of the Association
  for Computational Linguistics}, pages 2214--2220, Florence, Italy.
  Association for Computational Linguistics.

\bibitem[{Gao et~al.(2018)Gao, Meyer, and Gurevych}]{gao-etal-2018-april}
Yang Gao, Christian~M. Meyer, and Iryna Gurevych. 2018.
\newblock \href {https://doi.org/10.18653/v1/D18-1445} {{APRIL}: Interactively
  learning to summarise by combining active preference learning and
  reinforcement learning}.
\newblock In \emph{Proceedings of the 2018 Conference on Empirical Methods in
  Natural Language Processing}, pages 4120--4130, Brussels, Belgium.
  Association for Computational Linguistics.

\bibitem[{Goodrich et~al.(2019)Goodrich, Rao, Liu, and
  Saleh}]{goodrich2019assessing}
Ben Goodrich, Vinay Rao, Peter~J Liu, and Mohammad Saleh. 2019.
\newblock Assessing the factual accuracy of generated text.
\newblock In \emph{Proceedings of the 25th ACM SIGKDD International Conference
  on Knowledge Discovery \& Data Mining}, pages 166--175.

\bibitem[{Hokamp and Liu(2017)}]{hokamp-liu-2017-lexically}
Chris Hokamp and Qun Liu. 2017.
\newblock \href {https://doi.org/10.18653/v1/P17-1141} {Lexically constrained
  decoding for sequence generation using grid beam search}.
\newblock In \emph{Proceedings of the 55th Annual Meeting of the Association
  for Computational Linguistics (Volume 1: Long Papers)}, pages 1535--1546,
  Vancouver, Canada. Association for Computational Linguistics.

\bibitem[{Honnibal and Montani(2017)}]{spacy2}
Matthew Honnibal and Ines Montani. 2017.
\newblock {spaCy 2}: Natural language understanding with {B}loom embeddings,
  convolutional neural networks and incremental parsing.
\newblock To appear.

\bibitem[{Hu et~al.(2019{\natexlab{a}})Hu, Khayrallah, Culkin, Xia, Chen, Post,
  and Van~Durme}]{hu2019improved}
J~Edward Hu, Huda Khayrallah, Ryan Culkin, Patrick Xia, Tongfei Chen, Matt
  Post, and Benjamin Van~Durme. 2019{\natexlab{a}}.
\newblock Improved lexically constrained decoding for translation and
  monolingual rewriting.
\newblock In \emph{Proceedings of the 2019 Conference of the North American
  Chapter of the Association for Computational Linguistics: Human Language
  Technologies, Volume 1 (Long and Short Papers)}, pages 839--850.

\bibitem[{Hu et~al.(2019{\natexlab{b}})Hu, Khayrallah, Culkin, Xia, Chen, Post,
  and Van~Durme}]{hu-etal-2019-improved}
J.~Edward Hu, Huda Khayrallah, Ryan Culkin, Patrick Xia, Tongfei Chen, Matt
  Post, and Benjamin Van~Durme. 2019{\natexlab{b}}.
\newblock \href {https://doi.org/10.18653/v1/N19-1090} {Improved lexically
  constrained decoding for translation and monolingual rewriting}.
\newblock In \emph{Proceedings of the 2019 Conference of the North {A}merican
  Chapter of the Association for Computational Linguistics: Human Language
  Technologies, Volume 1 (Long and Short Papers)}, pages 839--850, Minneapolis,
  Minnesota. Association for Computational Linguistics.

\bibitem[{Kryscinski et~al.(2020)Kryscinski, McCann, Xiong, and
  Socher}]{kryscinski-etal-2020-evaluating}
Wojciech Kryscinski, Bryan McCann, Caiming Xiong, and Richard Socher. 2020.
\newblock \href {https://doi.org/10.18653/v1/2020.emnlp-main.750} {Evaluating
  the factual consistency of abstractive text summarization}.
\newblock In \emph{Proceedings of the 2020 Conference on Empirical Methods in
  Natural Language Processing (EMNLP)}, pages 9332--9346, Online. Association
  for Computational Linguistics.

\bibitem[{Lewis et~al.(2019)Lewis, Liu, Goyal, Ghazvininejad, Mohamed, Levy,
  Stoyanov, and Zettlemoyer}]{lewis2019bart}
Mike Lewis, Yinhan Liu, Naman Goyal, Marjan Ghazvininejad, Abdelrahman Mohamed,
  Omer Levy, Ves Stoyanov, and Luke Zettlemoyer. 2019.
\newblock Bart: Denoising sequence-to-sequence pre-training for natural
  language generation, translation, and comprehension.
\newblock \emph{arXiv preprint arXiv:1910.13461}.

\bibitem[{Lin(2004)}]{lin-2004-rouge}
Chin-Yew Lin. 2004.
\newblock \href {https://aclanthology.org/W04-1013} {{ROUGE}: A package for
  automatic evaluation of summaries}.
\newblock In \emph{Text Summarization Branches Out}, pages 74--81, Barcelona,
  Spain. Association for Computational Linguistics.

\bibitem[{Liu and Lapata(2019)}]{liu-lapata-2019-text}
Yang Liu and Mirella Lapata. 2019.
\newblock \href {https://doi.org/10.18653/v1/D19-1387} {Text summarization with
  pretrained encoders}.
\newblock In \emph{Proceedings of the 2019 Conference on Empirical Methods in
  Natural Language Processing and the 9th International Joint Conference on
  Natural Language Processing (EMNLP-IJCNLP)}, pages 3730--3740, Hong Kong,
  China. Association for Computational Linguistics.

\bibitem[{Matsumaru et~al.(2020)Matsumaru, Takase, and
  Okazaki}]{matsumaru-etal-2020-improving}
Kazuki Matsumaru, Sho Takase, and Naoaki Okazaki. 2020.
\newblock \href {https://doi.org/10.18653/v1/2020.acl-main.123} {Improving
  truthfulness of headline generation}.
\newblock In \emph{Proceedings of the 58th Annual Meeting of the Association
  for Computational Linguistics}, pages 1335--1346, Online. Association for
  Computational Linguistics.

\bibitem[{Maynez et~al.(2020)Maynez, Narayan, Bohnet, and
  McDonald}]{maynez-etal-2020-faithfulness}
Joshua Maynez, Shashi Narayan, Bernd Bohnet, and Ryan McDonald. 2020.
\newblock \href {https://doi.org/10.18653/v1/2020.acl-main.173} {On
  faithfulness and factuality in abstractive summarization}.
\newblock In \emph{Proceedings of the 58th Annual Meeting of the Association
  for Computational Linguistics}, pages 1906--1919, Online. Association for
  Computational Linguistics.

\bibitem[{Nallapati et~al.(2016)Nallapati, Zhou, dos Santos,
  G{\"{u}}l{\c{c}}ehre, and Xiang}]{nallapati-etal-2016-abstractive-new}
Ramesh Nallapati, Bowen Zhou, C{\'{\i}}cero~Nogueira dos Santos, {\c{C}}aglar
  G{\"{u}}l{\c{c}}ehre, and Bing Xiang. 2016.
\newblock \href {https://doi.org/10.18653/v1/k16-1028} {Abstractive text
  summarization using sequence-to-sequence rnns and beyond}.
\newblock In \emph{Proceedings of the 20th {SIGNLL} Conference on Computational
  Natural Language Learning, CoNLL 2016, Berlin, Germany, August 11-12, 2016},
  pages 280--290. {ACL}.

\bibitem[{Nan et~al.(2021)Nan, Nallapati, Wang, Nogueira~dos Santos, Zhu,
  Zhang, McKeown, and Xiang}]{nan-etal-2021-entity}
Feng Nan, Ramesh Nallapati, Zhiguo Wang, Cicero Nogueira~dos Santos, Henghui
  Zhu, Dejiao Zhang, Kathleen McKeown, and Bing Xiang. 2021.
\newblock \href {https://aclanthology.org/2021.eacl-main.235} {Entity-level
  factual consistency of abstractive text summarization}.
\newblock In \emph{Proceedings of the 16th Conference of the European Chapter
  of the Association for Computational Linguistics: Main Volume}, pages
  2727--2733, Online. Association for Computational Linguistics.

\bibitem[{Narayan et~al.(2018)Narayan, Cohen, and
  Lapata}]{narayan-etal-2018-dont}
Shashi Narayan, Shay~B. Cohen, and Mirella Lapata. 2018.
\newblock \href {https://doi.org/10.18653/v1/D18-1206} {Don{'}t give me the
  details, just the summary! topic-aware convolutional neural networks for
  extreme summarization}.
\newblock In \emph{Proceedings of the 2018 Conference on Empirical Methods in
  Natural Language Processing}, pages 1797--1807, Brussels, Belgium.
  Association for Computational Linguistics.

\bibitem[{Narayan et~al.(2021)Narayan, Zhao, Maynez, Simoes, and
  McDonald}]{narayan2021planning}
Shashi Narayan, Yao Zhao, Joshua Maynez, Gon{\c{c}}alo Simoes, and Ryan
  McDonald. 2021.
\newblock Planning with entity chains for abstractive summarization.
\newblock \emph{arXiv preprint arXiv:2104.07606}.

\bibitem[{Post and Vilar(2018)}]{post-vilar-2018-fast}
Matt Post and David Vilar. 2018.
\newblock \href {https://doi.org/10.18653/v1/N18-1119} {Fast lexically
  constrained decoding with dynamic beam allocation for neural machine
  translation}.
\newblock In \emph{Proceedings of the 2018 Conference of the North {A}merican
  Chapter of the Association for Computational Linguistics: Human Language
  Technologies, Volume 1 (Long Papers)}, pages 1314--1324, New Orleans,
  Louisiana. Association for Computational Linguistics.

\bibitem[{See et~al.(2017)See, Liu, and Manning}]{see-etal-2017-get}
Abigail See, Peter~J. Liu, and Christopher~D. Manning. 2017.
\newblock \href {https://doi.org/10.18653/v1/P17-1099} {Get to the point:
  Summarization with pointer-generator networks}.
\newblock In \emph{Proceedings of the 55th Annual Meeting of the Association
  for Computational Linguistics (Volume 1: Long Papers)}, pages 1073--1083,
  Vancouver, Canada. Association for Computational Linguistics.

\bibitem[{Shapira et~al.(2020)Shapira, Pasunuru, Ronen, Bansal, Amsterdamer,
  and Dagan}]{shapira2020evaluating}
Ori Shapira, Ramakanth Pasunuru, Hadar Ronen, Mohit Bansal, Yael Amsterdamer,
  and Ido Dagan. 2020.
\newblock Evaluating interactive summarization: an expansion-based framework.
\newblock \emph{arXiv preprint arXiv:2009.08380}.

\bibitem[{Sun et~al.(2020)Sun, Xiong, Liu, Liu, and Bao}]{sun2020joint}
Si~Sun, Chenyan Xiong, Zhenghao Liu, Zhiyuan Liu, and Jie Bao. 2020.
\newblock Joint keyphrase chunking and salience ranking with bert.
\newblock \emph{arXiv preprint arXiv:2004.13639}.

\bibitem[{Zhang et~al.(2020{\natexlab{a}})Zhang, Zhao, Saleh, and
  Liu}]{zhang2019pegasus}
Jingqing Zhang, Yao Zhao, Mohammad Saleh, and Peter~J Liu. 2020{\natexlab{a}}.
\newblock Pegasus: Pre-training with extracted gap-sentences for abstractive
  summarization.
\newblock In \emph{ICML}.

\bibitem[{Zhang et~al.(2020{\natexlab{b}})Zhang, Wang, Li, Gan, Brockett, and
  Dolan}]{zhang-etal-2020-pointer}
Yizhe Zhang, Guoyin Wang, Chunyuan Li, Zhe Gan, Chris Brockett, and Bill Dolan.
  2020{\natexlab{b}}.
\newblock \href {https://doi.org/10.18653/v1/2020.emnlp-main.698} {{POINTER}:
  Constrained progressive text generation via insertion-based generative
  pre-training}.
\newblock In \emph{Proceedings of the 2020 Conference on Empirical Methods in
  Natural Language Processing (EMNLP)}, pages 8649--8670, Online. Association
  for Computational Linguistics.

\bibitem[{Zhu et~al.(2021)Zhu, Hinthorn, Xu, Zeng, Zeng, Huang, and
  Jiang}]{zhu-etal-2021-enhancing}
Chenguang Zhu, William Hinthorn, Ruochen Xu, Qingkai Zeng, Michael Zeng,
  Xuedong Huang, and Meng Jiang. 2021.
\newblock \href {https://doi.org/10.18653/v1/2021.naacl-main.58} {Enhancing
  factual consistency of abstractive summarization}.
\newblock In \emph{Proceedings of the 2021 Conference of the North American
  Chapter of the Association for Computational Linguistics: Human Language
  Technologies}, pages 718--733, Online. Association for Computational
  Linguistics.

\end{thebibliography}
\bibliographystyle{acl_natbib}

\newpage
\clearpage
\appendix

\section{Human Evaluation}
\label{sec:human}

We compare the constrained summary $s$ and unconstrained summary $s'$ produced by the same base model for an apple-to-apple comparison.
We give the annotators the following guidelines: CAS is considered as better if it corrects factual inconsistencies in $s'$ or provides additional information supported by the reference summary. 
CAS is considered as worse if it leads to unfaithful or unsmooth summary.
CAS is considered similar to the unconstrained summary for other cases, most of which are as follows: (Both good) when the semantics of $s'$ is related to the reference summary, $s$ may rephrase parts of $s'$ or adds (drops) minor facts without changing the overall semantics.
(Both bad): when $s'$ is off-topic, it is most likely that $s$ is irrelevant as well.
\textbf{We provide examples of all cases mentioned above in App.~\ref{sec:examples}}.

We note that it is hard to evaluate factual consistency via crowdsourcing as the inter-annotator agreement and general quality of crowdsourcing annotations for factual consistency tend to be too low to be considered reliable \cite{kryscinski-etal-2020-evaluating} and many previous human evaluations on factual consistency, including ours, involve expert annotators \cite{kryscinski-etal-2020-evaluating,matsumaru-etal-2020-improving,nan-etal-2021-entity}. 
The Fleiss’s Kappa for our 3 annotators (including two non-authors) is 0.45. Human evaluations of similar sizes (50/100/109/10 samples) are conducted by previous studies \cite{dong-etal-2020-multi,zhu-etal-2021-enhancing,matsumaru-etal-2020-improving,nan-etal-2021-entity}.

\section{Effectiveness of Constraint Guidance}
\label{app:beam}
As a supplement to Fig.~\ref{fig:beam}, we list the detailed results of CAS when different beam sizes are used in Tables~\ref{tab_xsum_beam} and \ref{tab_cnndm_beam}.
The observations on XSum and CNNDM are consistent -- larger beam size leads to better performance for CAS.
When varying the beam size of unconstrained generation, the performance changes are negligible and thus unlisted.

\begin{table}[ht]
\centering
\scalebox{.85}{
\begin{tabular}{lrrrr}
\toprule
\multirow{2}{*}{\textbf{Constraint Type}}& \multicolumn{4}{c}{\textbf{XSum}}\\
 & \textbf{R-1} & \textbf{R-2} & \textbf{R-L} & \textbf{B}\\
\midrule
None &38.91	&16.54	&31.3 &5\\

\midrule
Entity $\wedge$ miss &         46.87  &         21.55  &         33.92  & 5 \\
Entity $\wedge$ miss &         47.83  &         23.00  &         36.13  & 10 \\
Entity $\wedge$ miss &         48.18  &         23.64  &         36.86  & 20 \\
Entity $\wedge$ miss &         \textbf{48.46}  &         \textbf{24.17}  &         \textbf{37.47}  & 50 \\
\midrule
Phrase-4      &         47.56  &         27.71  &         38.72  & 5 \\
Phrase-4      &         49.29  &         29.64  &         41.78  & 20 \\
Phrase-4      & \textbf{49.76} & \textbf{30.33} & \textbf{42.55} & 50 \\

\bottomrule
\end{tabular}
}
\upv
\caption{\textbf{Performance changes by beam size on the XSum dataset}.  B denotes the beam size.
}
\label{tab_xsum_beam}
\end{table}
\begin{table}[ht]
\centering

\scalebox{.85}{
\begin{tabular}{lrrrr}
\toprule
\multirow{2}{*}{\textbf{Constraint Type}}& \multicolumn{4}{c}{\textbf{CNNDM}}\\
 & \textbf{R-1} & \textbf{R-2} & \textbf{R-L} & \textbf{B}\\
\midrule
None &42.00	&19.44	&38.98 & 5\\

\midrule
Entity  &43.31 &19.57 &40.05 &5  \\
Entity  &         \textbf{44.75}  &         \textbf{21.34}  &         \textbf{41.60} &20 \\
\midrule
Phrase-4 &         43.37  &         21.57  &         40.82 &5 \\
Phrase-4 &         44.77  &         22.80  &         42.14 &10 \\
Phrase-4 & \textbf{45.14} & \textbf{23.21} & \textbf{42.43}&20 \\

\bottomrule
\end{tabular}
}
\upv
\caption{\textbf{Performance changes by beam size on the CNNDM dataset}.  B denotes the beam size.
}
\label{tab_cnndm_beam}
\downv
\end{table}

\begin{table*}[ht]
    \centering
    \scalebox{.88}{
    \begin{tabular}{p{1.15\linewidth}}
        \toprule

\textbf{Remarks}: CAS is effective at correcting entity information such as locations. \colorG{inconsistent \textrightarrow consistent (better) \checkmark}\\
\textbf{Constraint Set}:  ['cardiff'] \\
\textbf{Reference}:  hundreds of green-fanged tube web spiders have taken over the back garden of a family home in \colorG{cardiff} . \\
\textbf{Unconstrained}:  a man from \colorR{carmarthenshire} has appealed for help to find a `` extremely rare '' colony of tube spiders . \\
\textbf{Constrained}: \ \ \ \  a man from \colorG{cardiff} has said he is concerned about the number of `` extremely rare '' tube spiders . \\
\midrule

\textbf{Remarks}: CAS successfully distinguishes and replaces one of the two entities (``a'' and ``two'') of the cardinal type in the unconstrained summary. \colorG{inconsistent \textrightarrow consistent (better) \checkmark}\\
\textbf{Constraint Set}:  ['three'] \\
\textbf{Reference}:  two men have been assaulted by \colorG{three} masked attackers at an address in the craigmillar area of edinburgh . \\
\textbf{Unconstrained}:  a man is in a critical condition in hospital after being attacked by \colorR{two} men at a house in glasgow . \\
\textbf{Constrained}: \ \ \ \  a man is in a critical condition in hospital after being attacked by \colorG{three} men at a house in glasgow . \\
\midrule

\textbf{Remarks}: CAS adds the date ``1991'' at a proper position. \colorG{consistent \textrightarrow consistent (better) \checkmark}\\
\textbf{Constraint Set}:  ['1991'] \\
\textbf{Reference}:  toddler ben needham `` most likely '' died in an accident near to where he disappeared \colorG{in 1991} , police have said . \\
\textbf{Unconstrained}:  police investigating the disappearance of missing toddler ben needham have closed off a `` large number of theories '' about what happened to him . \\
\textbf{Constrained}: \ \ \ \  police investigating the disappearance of missing toddler ben needham \colorG{in 1991} have closed off a `` large number '' of theories about what happened to him . \\
\midrule

\textbf{Remarks}: CAS adds not only the constraint ``greece'' but two other entities appearing in the reference summary (``london'' and ``saturday'') to the output. However, it omits ``eurobasket warm-up game'' that was covered by the unconstrained summary. \colorG{consistent \textrightarrow consistent (tie) \checkmark}\\
\textbf{Constraint Set}:  ['greece'] \\
\textbf{Reference}:  bbc sport is showing live coverage of the eurobasket warm-up game between great britain and \colorG{greece} at the copper box in london on saturday 19 august . \\
\textbf{Unconstrained}:  great britain 's men will be shown live on television for the first time in their \colorG{eurobasket 2017 warm-up game} . \\
\textbf{Constrained}: \ \ \ \  great britain 's men will be shown live on television for the first time when they face \colorG{greece} in \colorG{london} on \colorG{saturday} . \\
\midrule

\textbf{Remarks}: CAS replaces ``argentine'' with ``argentina 's'' due to the provided constraint and the remaining of the sentence is unchanged. \colorG{consistent \textrightarrow consistent (tie) \checkmark}\\
\textbf{Constraint Set}:  ['argentina'] \\
\textbf{Reference}:  argentine president cristina fernandez and amnesty international have called for justice after the violent death of a transgender activist . \\
\textbf{Unconstrained}:  \colorG{argentine} president cristina fernandez de kirchner has called for an investigation into the murder of a transgender woman . \\
\textbf{Constrained}: \ \ \ \  \colorG{argentina} 's president cristina fernandez de kirchner has called for an investigation into the murder of a transgender woman . \\
        
        \bottomrule
    \end{tabular}
    }
    \upv
    \caption{\textbf{Examples showing good cases of CAS when constraints are automatically extracted}.} 
    \downv
    \label{tab_examples_good}
    
\end{table*}

\section{More Examples of CAS}
\label{sec:examples}
\start{Automatic Summarization}
In Tables~\ref{tab_examples_good} and~\ref{tab_examples_bad}, we show examples of CAS in automatic summarization where the constraints are extracted from the source documents. The examples with remarks cover both good and bad cases: inconsistent \textrightarrow consistent, consistent \textrightarrow consistent, consistent \textrightarrow inconsistent, and inconsistent \textrightarrow inconsistent.
We observe that CAS can correct factual inconsistencies or add relevant facts when the semantics of the unconstrained summary is close to the reference summary. On the other hand, CAS is not enough to change the overall message when the system summary is too far away from the reference, which is somewhat expected as CAS only affects model decoding without additional training.
Such observations suggest that one could possibly obtain better results by using stronger base models.

\begin{table*}[ht]
    \scalebox{1}{
    \begin{tabular}{p{1\linewidth}}
        \toprule

\textbf{Remarks}: CAS changes ``an american'' to ``two americans'' as the extracted constraint itself is wrong. \colorR{consistent \textrightarrow inconsistent (worse) \texttimes}\\
\textbf{Constraint Set}:  ['americans'] \\
\textbf{Reference}:  a french-american man who helped stop a heavily armed gunman on a train in france in 2015 has received the country 's highest honour . \\
\textbf{Unconstrained}:  an american man has been awarded france 's highest honour for his role in a terror attack on a train . \\
\textbf{Constrained}: \ \ \ \  \colorR{two americans} have been awarded the legion d'honneur , the highest honour in france for the paris terror attack . \\
\midrule

\textbf{Remarks}: CAS fails to recognize ``camilla'' is a synonym of ``duchess of cornwall'' and assigns the name to their son by mistake. \colorR{consistent \textrightarrow inconsistent (worse) \texttimes}\\
\textbf{Constraint Set}:  ['camilla'] \\
\textbf{Reference}:  prince charles and his wife camilla are to visit australia and new zealand in november , his office says . \\
\textbf{Unconstrained}:  prince charles and duchess of cornwall will travel to australia for the first time since the start of the year . \\
\textbf{Constrained}: \ \ \ \  prince charles and duchess of cornwall will travel to australia for the first time since \colorR{their son camilla} was born in november . \\
\midrule

\textbf{Remarks}: CAS at times leads to unsmooth output when it fails to recognize the constraints (``royal yacht britannia'' in this example), although such cases are rare. \colorR{consistent \textrightarrow inconsistent (worse) \texttimes}\\
\textbf{Constraint Set}:  ['britannia'] \\
\textbf{Reference}:  the royal yacht \colorG{britannia} has been judged as scotland 's best visitor attraction every year for the past decade . \\
\textbf{Unconstrained}:  the royal yacht has been named scotland 's best visitor attraction for 10 years , according to a new report . \\
\textbf{Constrained}: \ \ \ \  the royal yacht of scotland has been named the best visitor attraction in scotland for 10 years . \colorG{britannia} . \\
\midrule

\textbf{Remarks}: CAS fails as the semantics of the unconstrained summary is too far away from the reference. \colorR{inconsistent \textrightarrow inconsistent (tie) \texttimes}\\
\textbf{Constraint Set}:  ['twitter'] \\
\textbf{Reference}:  a tweet from one direction 's louis tomlinson to harry styles has become the second most retweeted post of all time . \\
\textbf{Unconstrained}:  louis harris has become \colorR{the most popular person in the history of one direction} and has been retweeted . \\
\textbf{Constrained}: \ \ \ \  louis z has become \colorR{the most popular twitter post in the us} - and he 's never in his heart . \\
\midrule

\textbf{Remarks}: CAS is ineffective when the unconstrained summary is unsmooth or off-topic, even if the quality of the constraints is high. \colorR{inconsistent \textrightarrow inconsistent (tie) \texttimes}\\
\textbf{Constraint Set}:  ['harry potter'] \\
\textbf{Reference}:  the author of the \colorG{harry potter} books , jk rowling , is celebrating her 50th birthday today . \\
\textbf{Unconstrained}:  the official cake of \colorR{jk rowling 's jk rowling cake} has been released to mark the \colorR{90th anniversary} of the birth of the author . \\
\textbf{Constrained}: \ \ \ \  a cake made by jk rowling has been unveiled to mark the \colorR{70th anniversary} of jk rowling 's \colorG{harry potter} cake . \\

        \bottomrule
    \end{tabular}
    }
    \caption{\textbf{Examples showing bad cases of CAS when constraints are automatically extracted}.} 
    \label{tab_examples_bad}
    
\end{table*}

\start{Interactive Summarization}
In Table~\ref{tab_examples}, we show the examples of CAS for interactive summarization, which are \textit{randomly sampled} from the XSum dataset. The performance of CAS in interactive summarization is generally better than automatic summarization as the constraints are manually provided, which shows the benefits of human feedback as well as the potential of CAS when constraints of higher quality are available.

\begin{table*}[ht]
    \scalebox{.96}{
    \begin{tabular}{p{1\linewidth}}
        \toprule

\textbf{Constraint Set}:  ['three'] \colorG{\checkmark}\\
\textbf{Reference}:  \colorG{three} men have been found guilty of killing a rival drug dealer in a gang-related revenge attack . \\
\textbf{Unconstrained}:  \colorR{four} men have been convicted of killing a father-of-two in a `` vicious '' attack in a denbighshire car park . \\
\textbf{Constrained}: \ \ \ \  \colorG{three} men have been convicted of killing a father-of-two in a `` vicious '' attack in a denbighshire car park . \\
\midrule
\textbf{Constraint Set}:  ['monetary policy committee'] \colorG{\checkmark}\\
\textbf{Reference}:  uk interest rates have been held at 0.5 \% again by the bank of england 's \colorG{monetary policy committee} -lrb- mpc -rrb- . \\
\textbf{Unconstrained}:  the bank of england -lrb- \colorR{mpc} -rrb- has voted to keep interest rates at its current rate for the first time this year . \\
\textbf{Constrained}: \ \ \ \  the bank of england \colorG{monetary policy committee} -lrb- mpc -rrb- has voted to keep interest rates at its current rate . \\
\midrule
\textbf{Constraint Set}:  ['portuguese', 'tuesday'] \colorG{\checkmark}\\
\textbf{Reference}:  six people died and another two were missing after an explosion on \colorG{tuesday} evening destroyed a fireworks factory near the \colorG{portuguese} town of lamego . \\
\textbf{Unconstrained}:  the authorities in the \colorR{brazilian} city of porto have widened their search for eight people killed in an explosion at a factory . \\
\textbf{Constrained}: \ \ \ \  the authorities in the \colorR{eastern} \colorG{portuguese} city of porto have widened their search for eight people killed in \colorG{tuesday} 's explosion at a factory . \\
\midrule
\textbf{Constraint Set}:  ['wales', 'public health wales'] \colorR{\texttimes}\\
\textbf{Reference}:  an outbreak of 189 cases of measles has been reported in swansea , neath and port talbot , \colorG{public health wales} says . \\
\textbf{Unconstrained}:  the number of school children caught measles in swansea and neath port talbot has reached a record high . \\
\textbf{Constrained}: \ \ \ \  \colorG{public health wales} has said it is concerned at the number of cases of measles in \colorG{wales} . \\
\midrule
\textbf{Constraint Set}:  ['sicily', 'cesare'] \textendash\\
\textbf{Reference}:  from the terrace of his winery near the baroque town of caltagirone in south-eastern \colorG{sicily} , \colorG{cesare} nicodemo surveys his fields of ripening vines - a glass of his finest spumante in hand . \\
\textbf{Unconstrained}:  it 's almost a year since the sicilian city of mauricio pounced its name to the rural mafia , but it 's not quite like it . \\
\textbf{Constrained}: \ \ \ \  it 's almost a year since \colorG{sicily} 's \colorG{cesare} deglazio took on the streets of rome . \\
\midrule
\textbf{Constraint Set}:  ['northern ireland'] \textendash\\
\textbf{Reference}:  the leader of the catholic church in ireland has backed a call by amnesty international for an inquiry into mother-and-baby homes in \colorG{northern ireland} . \\
\textbf{Unconstrained}:  the archbishop of canterbury justin martin has called for an investigation into the transfer of human remains from the catholic church . \\
\textbf{Constrained}: \ \ \ \  the archbishop of canterbury justin martin has called for an immediate investigation into the transfer of human remains to \colorG{northern ireland} . \\
\midrule
\textbf{Constraint Set}:  ['xinjiang'] \colorG{\checkmark}\\
\textbf{Reference}:  china has executed eight people in the north-western region of \colorG{xinjiang} , for what it calls `` terrorist '' attacks , reports the state news agency xinhua . \\
\textbf{Unconstrained}:  china 's state-run news agency xinhua has executed three men for terrorism , state media report . \\
\textbf{Constrained}: \ \ \ \  china 's state-run news agency xinhua says eight men convicted of terrorism offences in \colorG{xinjiang} have been executed . \\

        \bottomrule
    \end{tabular}
    }
    \caption{\textbf{\textit{Randomly sampled} examples of CAS in interactive summarization}. We filter the entity constraints such that they are not in the unconstrained summary (Entity $\wedge$ miss) to examine the effectiveness of CAS for post-editing.} 
    \label{tab_examples}
    
\end{table*}

\end{document}